\documentclass[10pt,twocolumn,letterpaper]{article}

\usepackage[pagenumbers]{cvpr}              %
\usepackage{graphicx}
\usepackage{amsmath}
\usepackage{amssymb}
\usepackage[outdir=./]{epstopdf}

\usepackage{scalerel}
\usepackage{placeins}
\usepackage{booktabs}
\usepackage{xcolor}         %
\usepackage{xspace}
\usepackage{enumitem}
\usepackage{adjustbox}
\usepackage[pagebackref,breaklinks,colorlinks]{hyperref}

\usepackage[capitalize]{cleveref}
\crefname{section}{Sec.}{Secs.}
\Crefname{section}{Section}{Sections}
\Crefname{table}{Table}{Tables}
\crefname{table}{Tab.}{Tabs.}

\definecolor{bppgapcol}{RGB}{45,162,170}
\definecolor{red}{RGB}{192,55,44}

\usepackage{soul}

\def\cvprPaperID{6689} %
\def\confName{CVPR}
\def\confYear{2023}

\begin{document}

\title{HFLIC: Human Friendly Perceptual  Learned  \\
Image Compression with Reinforced Transform}%

\author{Peirong Ning\\
Shenzhen Graduate School, \\
Peking University \\
{\tt\small beiluo97@gmail.com}
\and
Wei Jiang\\
Shenzhen Graduate School,\\
Peking University\\
{\tt\small wei.jiang1999@outlook.com }%
\and
Ronggang Wang\\
Shenzhen Graduate School,\\
Peking University\\
{\tt\small rgwang@pkusz.edu.cn }%
}
\maketitle
\begin{abstract}
In recent years, there has been rapid development in learned image compression techniques that prioritize rate-distortion-perceptual compression, preserving fine details even at lower bit-rates. However, current learning-based image compression methods often sacrifice human-friendly compression and require long decoding times. In this paper, we propose enhancements to the backbone network and loss function of existing image compression model, focusing on improving human perception and efficiency. Our proposed approach achieves competitive subjective results compared to state-of-the-art end-to-end learned image compression methods and classic methods, while requiring less decoding time and offering human-friendly compression. Through empirical evaluation, we demonstrate the effectiveness of our proposed method in achieving outstanding performance, with more than $25\%$ bit-rate saving with comparable perceptual quality.

\end{abstract}
\section{Introduction}
Over the last few years, reducing bit rates while maintaining detail has become increasingly challenging. This concept is formalized in the fundamental rate-distortion trade-off, where "rate" refers to bit rate, and "distortion" refers to the pairwise comparison between the input image and the reconstruction. Classical image compression standards include JPEG~\cite{jpeg1992wallace1}, BPG ~\cite{bpgurl2}, and progressing Versatile Video Coding (VVC) \cite{vtm173},  which try to minimize this trade-off. These years, learned image compression 
(LIC), \cite{jiang2023slic31}, \cite{he2022elic4} and \cite{jiang2022mlic5} has outperformed traditional methods, base on variation auto-encoders with analysis transform, synthesis transform and entropy model. However, purely rate-distortion optimized systems will produce artifacts in the reconstruct images both in traditional hand-craft method or learning based neural approaches. An increasing amount of interest has been given to examining the reconstruction "realistically" or "perceptually quality". 

Existing methods
~\cite{po-elic6} have accomplish pretty good performance, \cite{mentzer2020high7} and \cite{agustsson2019extreme8}  introduce generative adversarial network (GAN) to enhance realism in hyperprior-based compression architecture. Additionally, \cite{ma2021variable34}, \cite{ma2021afec9} and \cite{li2022content10} introduce importance maps and region-of-interest (ROI) masks to allocate more bits to important areas, or combine multiple loss terms during the training phase. In \cite{agustsson2022multi32},a method which is capable of outputting a single representation for compressed images, from which a receiver can either decode a high-realism reconstruction or a high-PSNR reconstruction. And \cite{muckley2023improving33} introduce a non-binary discriminator that is conditioned on quantized local image representations obtained via  VQ-VAE autoencoders.
However, these method haven't combine the human friend perceptual and efficient decoding time.
\begin{figure*}[htb]
    \centering
    \includegraphics[width=0.98\linewidth]{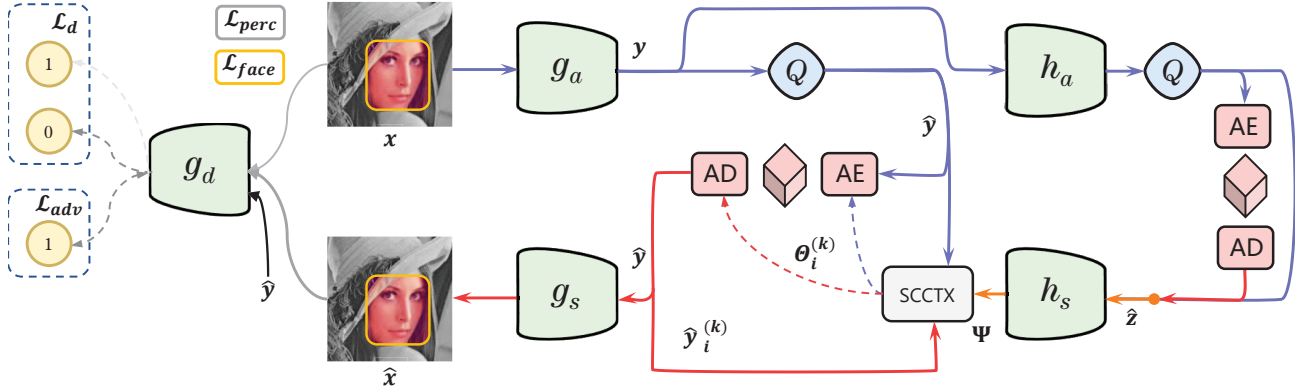}
    \caption{\label{label} Diagram of the adopted framework. The right part is our enhancement LIC. We use the same architecture of $g_a$, $g_s$, $h_a$ and $h_s$ as the original paper ELIC\cite{he2022elic4}. SCCTX denotes the spatial-channel context model. We use the uneven 5-group scheme with parallel context models \cite{he2021Checkerboard13}. The left
part shows the adversarial training. We use the same discriminator ($g_d$) structure as HiFiC~\cite{mentzer2020high7}.}
\end{figure*} 

Our goal is bridge the decoding efficient and reconstruction perceptual quality. Channel-wise auto-regressive entropy model\cite{Minnen20channel14} will improve compression so much, but it will bring more time into decoding part. For most of compression tasks, they will more care about decoding time rather than decoding time,  this inspire us to bring more expressive transforms to get compact latent to simplify the decoder. At the same time, humans are very sensitive to the areas where face parts appear, and distortion from perceptual loss will distort these parts, which requires us to optimize the face parts. Toward this goal, we explore two aspect enhancement : reinforced synthesis transform with inverted bottleneck block and content adaptive loss. The combined effect of these improvement is help us get a compact latent in and simplify the decoding process, and utilize lower bit-rate to achieve comparable visual quality against previous approaches. Our mse outperforms recently proposed model ELIC~\cite{he2022elic4} by 3.95\% on Kodak\cite{kodakurl15}, and save more than $20\%$ decoding time. Additionally, we get more than $25\%$ bitrate saving on qualitative compare with former methods.
\section{Reinforced synthesis transform with inverted bottleneck block}
Lossy image compression aims to optimize the rate distortion
function $ \mathcal{R}+ \lambda \mathcal{D}$. Denoting the image as $x$, encoder
as $g_a$ and decoder as $g_s$, the neural network has the following
objective:
\begin{equation}
\mathcal{L} = \mathbb{E}[-\log p(g_a(x)) + \lambda d (x, g_s(ga(x)))]
\end{equation}
where $\mathbb{E}$ is the expectation over $p(x)$,$g_a$ extracts the input image $x$ as latent variable $\hat{y} = g_a(x)$ and $g_s$ transforms it into reconstruction $\hat{x}$. 
$\mathcal{D}$, $\mathcal{R}$ are the reconstruction distortion loss and bit-rate computed via learned prior. Auto-regressive context model~\cite{minnen2018joint16} is the key factor to promote compression performance by more accurately modeling symbol probability. 
To be specific, the estimation of current symbol $y_i$ can leverage previous symbols 
$y_{< i}$:

\begin{equation}
p(y_i|y_{<i}) = p(y_i|\Phi y_{<i}))
\end{equation}
where $\Phi$ is context model of various form ~\cite{he2022elic4}, utilizes a spatial-channel context modelling, using decoded point as the the reference point, uneven grouping for channel-conditional(CC) adaptive coding. Since the later groups need to refer to the previous decoded channel groups, more group divisions will slower down the decoding time due to the sequential reference relationship. For this reason, we reinforced analysis and synthesis transform. Previous~\cite{he15Resnet17} inverted bottleneck in compression model method use bottleneck block as hidden dimension is 1/2 narrower than the input dimension from ResNet~\cite{he15Resnet17}, and spare the latent feature. One important design from transformer block is creates an inverted bottleneck, and the hidden dimension of MLP block is four times wider than input dimension. And this Transformer design is connected to the inverted bottleneck design with an expansion ratio of 4 used in ConvNets~\cite{liu22convnet18}. ConvNext also illustrate how it gets the decrease in runtime while flops increase. Inspired by this idea, we expand inverted bottleneck design with an expansion ratio of 2, which enlarge our hidden layer and efficiently encode the latent feature bottleneck to produce more compact latent features with less group divisions.
\footnote{Here \url{https://github.com/JiangWeibeta/ELIC} we borrow ELIC code, Enh-POELIC and HFLIC code is available at \url{https://github.com/beiluo97/HFLIC}.}
\begin{figure*}[htb]
    \centering
    \includegraphics{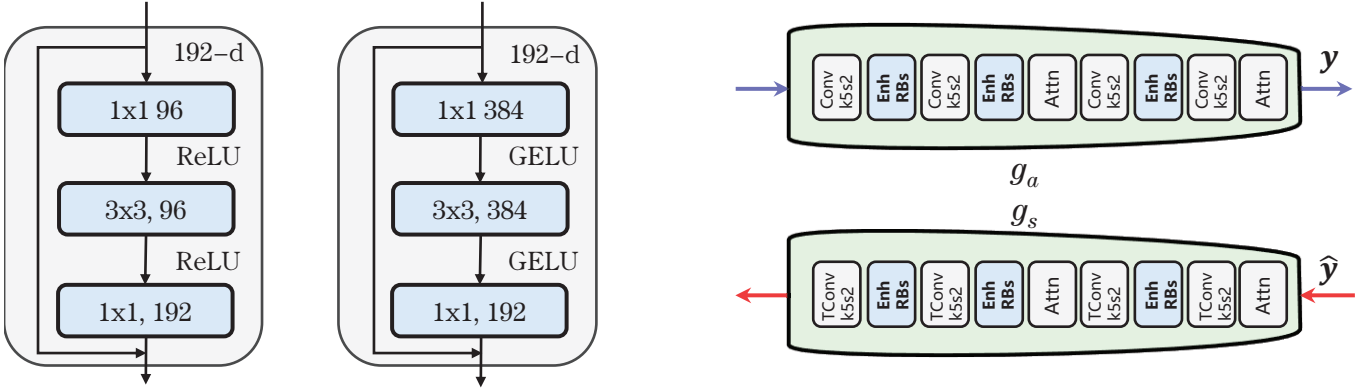}
    \caption{\label{label} Left one is the block designs for ResNet, which ELIC has adopted. Middle one is our, right one is architecture of $g_a$ amd $g_s$.  }
\end{figure*}

\section{Human Friendly Perceptual Loss} 
We take the rate-constrained RD optimization from HiFiC~\cite{mentzer2020high7} and POELIC~\cite{po-elic6}:
\begin{equation}
\mathcal{L} = \mathcal{D} + \lambda \cdot \mathcal{R}
\end{equation}
where $D$ and $R$ are (perceptual) distortion and rate terms. To focus on, Our  summarized perceptual $D$ loss function is 
\begin{equation}
\mathcal{D} =  M_{perc} \circ \mathcal{L}_{perc} +  M_{face} \circ \mathcal{L}_{face}  
\end{equation}
where the perceptual region loss $\mathcal{L}_{perc}$ contain a pixel-wise reconstruction loss $L_{recon}$ Charbonnier~\cite{Lai18chan19} loss, adversarial loss $\mathcal{L}_{adv}$~\cite{mentzer2020high7} ,perceptual loss LPIPS-VGG $L_{lpips}$~\cite{zhang18lpips20}, and the style loss $\mathcal{L}_{sty}$~\cite{Gatys17sty21}constraining the texture consistency. We will discuss these loss terms in detail in this section.

\subsection{Perceptual Texture and Structure Loss}
Texture and structure regions tend to have more details, and existing methods based on perceptual optimization [20] have achieved compelling results in texture reconstruction. Therefore, we use perceptual loss(chabonnier loss~\cite{Lai18chan19}, lpips~\cite{zhang18lpips20}, adversarial loss~\cite{mentzer2020high7}) in texture regions. 
 At the same time, perceptual loss functions such as LPIPS, which have a large receptive field, can introduce additional noise that is not acceptable for precise edge reconstruction. Hence, we combine patched style loss from ~\cite{po-elic6} with the existing texture loss method. For adversarial loss, we apply hinge loss to train a synthesizer with PatchGAN~\cite{demir18pgan22} discriminator.
 The perceptual texture and structure formulation can be expressed as follows:
\begin{equation}
\mathcal{L}_{perc} =   \omega_{rec}\mathcal{L}_{rec} + \omega_{lpips}\mathcal{L}_{lpips} + \omega_{adv}\mathcal{L}_{adv} + \omega_{sty}\mathcal{L}_{sty}   
\end{equation}
where $\omega_{rec}$, $\omega_{lpips}$, $\omega_{adv}$, $\omega_{sty}$ are weights of corresponding loss metrics. Note that we use VGG as lpips and style loss pretrained feature extraction network. 
\subsection{Human Friendly Small Face Loss}
The human eye is more sensitive to certain regions~\cite{li2022content10}, such as the face, especially small details. Therefore, strict constraints should be used to avoid deformation in these areas. In our framework, facial regions are classified as either texture regions or optimized with perceptual loss if left unaltered.
To address this issue, we adopt a different loss function for small faces. As people are particularly sensitive to the correctness of facial structure, accurate reconstruction is crucial. Therefore, we use a stricter constraint loss, the MSE loss, for facial image reconstruction.
\begin{equation}
\mathcal{L}_{face} =  \omega_{face} \cdot \mathcal{L}_{mse}  
\end{equation}
    where the Msface denotes the mask of the small face regions, and $\mathcal{L}_{mse}$ is the Mean Squared Error(MSE),  We use the well-known YOLO-v5-face~\cite{qi21yoloface23}to detect the faces in
the image, and $L_{face}$ is only adopted to small faces. The bitrate of the quantized latent representation $\hat{y}$ is estimated by the entropy module denoted by P,$R(\hat{y}) = -\log(P(\hat{y}))$. Finally, the total loss function of the whole image is summaried
as:
\begin{equation}
\mathcal{L}_{total} =   M_{perc} \circ \mathcal{L}_{perc}  +  M_{sface} \circ \mathcal{L}_{face}   + \lambda \cdot \mathcal{R}(\hat{y})
\end{equation}
We investigate a variety of distortion loss functions wight for all of then , and  select the most human friendly one. Besides, adding a pixel-level mask to MSE or MAE is easy with simple pointwise multiplication. But it is a little harder to used on LPIPS or GAN losses, because these two loss functions compute the feature losses and cannot correspond to mask pixel-to-pixel. Using the methods mentioned in the~\cite{li2022content10}, we give an appropriate result.
\section{Experiments}
\subsection{Datasets and metrics}
For training we utilize the train all of the picture bigger than 480x480 form CLIC profession\cite{clic202024}, DIV2k\cite{agu17div25}, Filckr 2K\cite{hsan18flickr26}, ImageNet\cite{deng09imagenet27} and coco17\cite{Lin14coco28}, here we finally have 61564 images for all models. We used the full-resolution versions of the images.  For evaluation, considering comparing with SOTA methods, we focus most of our results in the main body on CLIC2022\cite{clic202024} and kodak\cite{kodakurl15} because the first is the latest used by the neural image compression contest, later is always used to evaluation perceptual quality.

Our evaluation metrics are reference-based metrics, computed in the form $\rho(\hat{x},x)$, where $x$ is a ground-truth image and $\hat{x}$ 
is a compressed version of $x$ .
The handcrafted reference metrics of MS-SSIM and PSNR are standards for evaluate image compression methods. A drawback of optimization for the handcrafted metrics is that it can lead to blurring of the reconstructed images. For this reason, other reference metrics such as LPIPS\cite{zhang18lpips20}, it has been developed that more heavily favor preservation of texture and are more correlated with human judgment, but it is important to note that as reference metrics they can still trade off some statistical fidelity. Beside, IQT\cite{cheon21iqt30}and \cite{jiang2022image12}  that successfully applies a transformer architecture to a perceptual full-reference. This method combines a CNN backbone as a feature extractor, with a Transformer encoder-decoder to compare a reference and distorted images, and predict the quality score. 
\begin{figure*}[h]
\begin{minipage}{41.5pc}
    \centering
    \includegraphics[width=41.5pc]{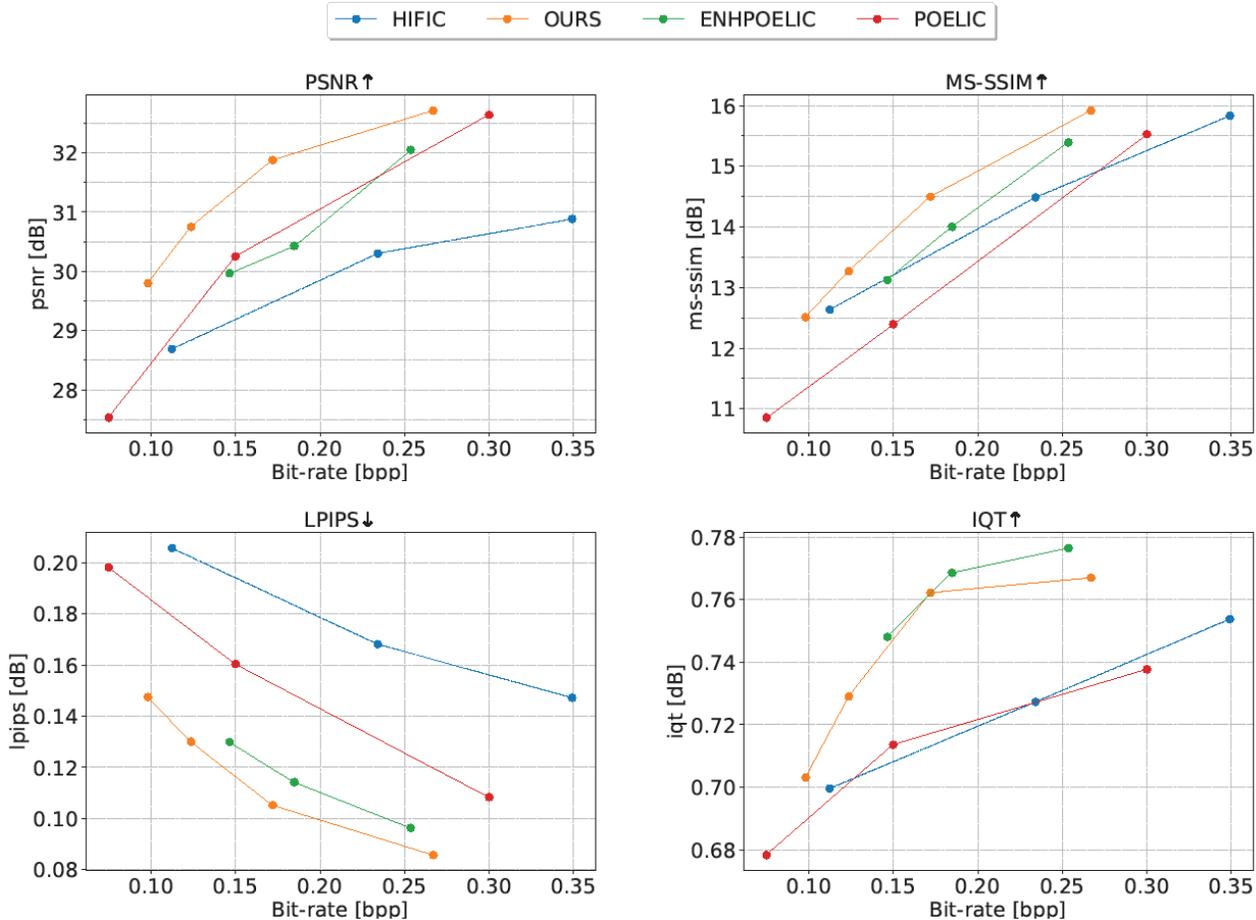}
    \caption{\label{label} Comparisons of methods across various distortion and statistical fidelity metrics for the CLIC 2022 validation set. Reference models (Enh-POELIC (Ours)) achieve the best IQT score, but display poor objective fidelity as measured by PSNR and MS-SSIM. Ours is able to achieve better subjective fidelity as measured by LPIPS\cite{zhang18lpips20} and IQT\cite{cheon21iqt30} vs. HiFiC at equivalent distortion levels..}
    \end{minipage}\hspace{2pc}%
\end{figure*} 
\begin{figure}[htb]
\includegraphics[width=8.3cm]{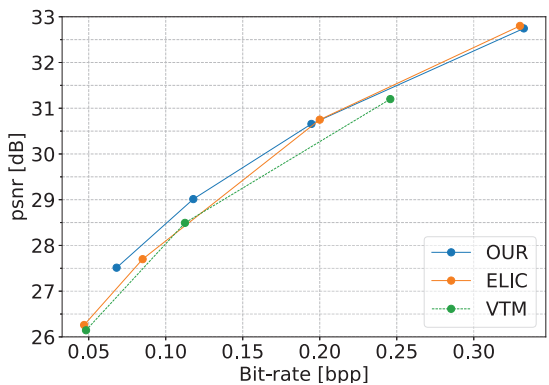}
\caption{RD and inference time of learned image compression models. The BD-Rate data is calculated relative to VVC (YUV 444) from
PSNR-BPP curve on Kodak.}
\end{figure}
\footnote{Implementations for fid, kid, dists, lpips, and iqt can be found at \url{https://github.com/beiluo97/ImageQualityAll}.}
\begin{table}[t]
  \centering
  \begin{tabular}{lccc}
    \toprule
    Model & BD-RATE & Enc (ms) & Dec (ms) \\
    \midrule
    ELIC   & -6.56\% & 259.4    & 240.6    \\
    ELIC-5 & -       & 195.5    & 154.8    \\
    OUR    & -9.87\% & 243.7    & 202.4    \\
    VTM    & 0       & -        & -        \\
    \bottomrule
  \end{tabular}
  \caption{Rate-distortion curves of ELIC(10-slice CC), ours(5-slice CC), and VTM\cite{vtm173}. The results are evaluated on Kodak. All shown learned models are optimized for minimizing MSE.}
\end{table}
\begin{figure*}[htb]
\begin{minipage}{41.3pc}
    \centering
    \includegraphics[width=41.3pc]{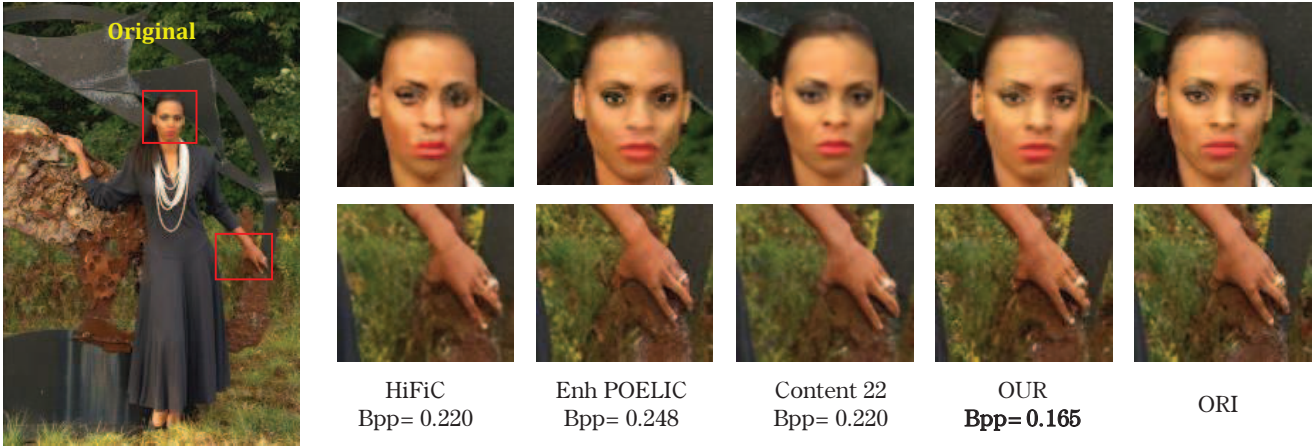}
    \caption{\label{label} Qualitative examples of compressed image kodim18. HiFiC and Enh POELIC optimized without $\mathcal{L}_{sface}$ shows heavily distortion in human face. OUR method shows we achieve more than $25\%$ bit-saving, and achieve better reconstruction performance.}
    \end{minipage}\hspace{2pc}%
\end{figure*} 
\begin{figure*}[htb]
\begin{minipage}{41.3pc}
    \centering
    \includegraphics[width=41.3pc]{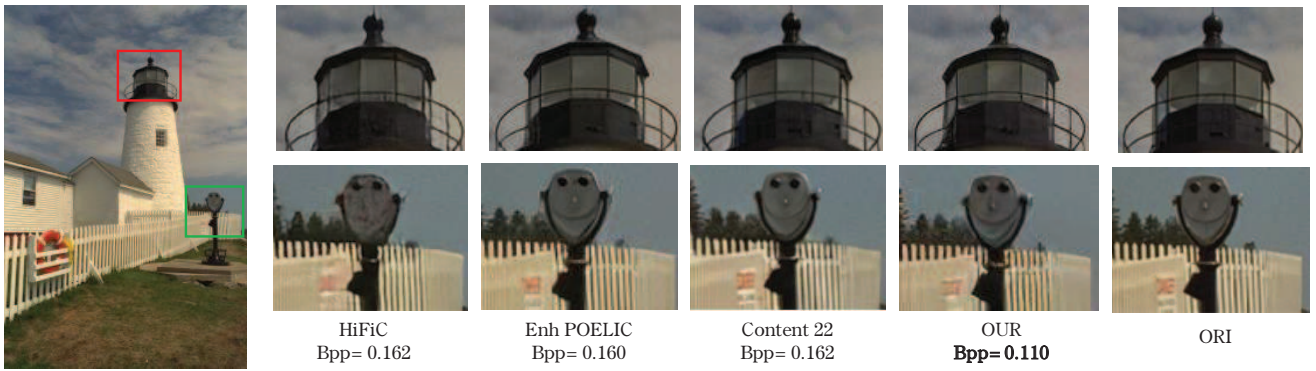}
    \caption{\label{label} Qualitative examples of compressed image kodim19. OUR method shows we achieve more than $25\%$ bit-saving, and achieve better reconstruction performance compare with former state-of-the-art model. }
    \end{minipage}\hspace{2pc}%
\end{figure*} 
\subsection{Baseline models}
To reveal the benefits of our enhancement transform while reducing time complexity, we refer to the training method in the ELIC and use MSE optimization to derive perceptually base models. In order to train low-rate models ($\lambda = \{8, 16, 32, 75\}\times 10^{-4})$, we train them with a $\lambda = 0.015$ at the beginning for 500 epochs, set the initial learning rate to 1e-4, set batch size to 8, and adjust them using target $\lambda$ values with learning rates of 1e-4 for 100 epochs, followed 3e-5, 1e-5, 3e-6, 1e-6 for 30 epochs each. We compare coding speed of our method with ELIC\cite{he2022elic4} and ELIC-5(5-slice CC), and the RD performance with ELIC and VTM-17.0. OUR method out perform ELIC and VTM on RD performance regarding PSNR. 
And we also follow \cite{po-elic6} use the same loss with our network, train Enh-POELIC for comparison.

\subsection{Quantitative results}
 Figure 3 illustrates the comparison of our method with other methods in terms of statistical fidelity metrics such as PSNR, MS-SSIM, LPIPS, and IQT. Across all bitrates, our method consistently outperforms previous work, exhibiting superior statistical fidelity.

\subsection{Qualitative results}

In order to assess the visual quality and fidelity of the reconstructed images, we conducted a qualitative comparison between enh-POELIC, HiFiC \cite{mentzer2020high7}, Content22 \cite{li2022content10}, and our proposed method. The experiments revealed that our method achieves higher fidelity even at lower bitrates.

Figure 5 and figure 6 showcases the qualitative comparison results.

\section{Conculusion}
In this study, we propose an enhanced transform approach for human-friendly Learned Image Compression (LIC). By combining an enhancement transform with a human-friendly loss, our neural architecture achieves pleasant reconstructions with fewer bits compared to previous methods. Perceptual metrics validate the high-fidelity nature of our approach. As future work, we aim to investigate the adaptability of the perceptual loss function.

\FloatBarrier
\newpage

{\small
\bibliographystyle{ieee_fullname}
\bibliography{egbib}
}

\end{document}